# Generalizing Case Frames Using a Thesaurus and the MDL Principle

Hang Li    Naoki Abe
C&C Research Laboratories, NEC
Miyazaki 4-1-1, Miyamae-ku, Kawasaki 216, Japan
{lihang,abe}@sbl.cl.nec.co.jp

## Abstract

We address the problem of automatically acquiring case frame patterns from large corpus data. In particular, we view this problem as the problem of estimating a (conditional) distribution over a partition of words, and propose a new generalization method based on the *MDL* (Minimum Description Length) principle. In order to assist with the efficiency, our method makes use of an existing thesaurus and restricts its attention on those partitions that are present as 'cuts' in the thesaurus tree, thus reducing the generalization problem to that of estimating the 'tree cut models' of the thesaurus. We then give an efficient algorithm which provably obtains the optimal tree cut model for the given frequency data, in the sense of MDL. We have used the case frame patterns obtained using our method to resolve pp-attachment ambiguity. Our experimental results indicate that our method improves upon or is at least as effective as existing methods.
**Keyword:** Corpus-Based Language Processing, Natural Language Learning, Case Frame, MDL Principle, Disambiguation

## 1 Introduction

We address the problem of automatically acquiring case frame patterns from large corpus data. A satisfactory solution of this problem would have a great impact on various tasks in natural language processing, such as the disambiguation problem in parsing, a central problem in this field. The acquired knowledge would also be helpful for building a lexicon, as it would provide lexicographers with word usage descriptions.

The purpose of the present research is to provide a method by which to acquire knowledge from limited data of observed case frames, which will allow us to judge the (degree of) acceptability of *unseen* case frames. Such an acquisition procedure will necessarily involve *generalization* of case frames available in the input data. The acquisition process will thus consist of two phases: *extraction* of case frame instances from corpus data, and *generalization* of those instances to case frame patterns. For the extraction problem, there have been various methods proposed to date, which are quite adequate (Brent 91; Hindle & Rooth 91; Grishman & Sterling 92; Manning 92; Smadja 93; Utsuro et al. 93). The generalization problem is a more challenging problem and has not been solved satisfactorily, although various methods have been proposed. Some of these methods make use of prior knowledge in the form of an existing thesaurus (Resnik 92; Resnik 93; Framis 94; Almuallim et al. 94), and others do not rely on any prior knowledge (Hindle 90; Brown et al. 92; Pereira et al. 93; Grishman & Sterling 94; Tanaka 94). In this paper, we propose a new generalization method which belongs to the first of these two categories.

We formalize the problem of generalizing case slots as that of estimating a model of probability distribution over some partition of words, and propose a new generalization method based on the *MDL* (Minimum Description Length) principle: a well-motivated and theoretically sound principle of statistical estimation from information theory. We also devised an efficient algorithm which is guaranteed to output an optimal model in the sense of *MDL*, provided we have a reliable thesaurus at hand. Finally we empirically tested the performance of our method, by using the generalized case frame patterns obtained by training our method with corpus data to resolve pp-attachment ambiguity. Our experimental results indicate that our method improves upon or is at least as effective as existing methods.

## 2 The Problem Setting

### 2.1 The Data Sparseness Problem

Suppose available to us are frequency data of the type shown in Figure 1, given by instances of a case frame automatically extracted from a corpus using conventional techniques. (In the sequel, we will refer to this type of frequency data as 'co-occurrence data.') The problem of generalizing case slots can be viewed as the problem of learning the underlying *conditional distribution* which gives rise to such data. Such a conditional distribution specifies the conditional probability[1]

$$P(n|v,s) \qquad (1)$$

for each $n$ in the set of nouns $\mathbf{N} = \{n_1, n_2, \ldots, n_N\}$, $v$ in the set of verbs $\mathbf{V} = \{v_1, v_2, \ldots, v_V\}$, and $s$ in the set of slot names $\mathbf{R} = \{s_1, s_2, \ldots, s_R\}$. (Such a probability model[2] is often referred to as a *word-based model*). Since the number of probability parameters in a word-based model is very large ($(N-1) \times V \times R$), a word-based model is difficult to estimate with a reasonable data size that is available in practice – a problem usually referred to as the 'data sparseness problem.' For example, suppose that we employ the well-known Maximum Likelihood Estimator (or MLE for short) to estimate the probability parameters of a word-based model given frequency data in Figure 1. MLE is obtained by simply *normalizing* the frequencies so that they sum to one, giving, for example, the estimated probabilities of 0.0, 0.2, and 0.4 for 'swallow,' 'eagle,' and 'bird,' respectively. Since in general the number of nouns exceeds the size of typically available data, MLE will result in estimating most of the probability parameters to be zero. To address this problem, Grishman & Sterling proposed a method of *smoothing* the probabilities using a similarity measure between words, which itself is calculated based on co-occurrence data (Grishman & Sterling 94). That is, probability estimates of words are smoothed by weighted averaging using the similarity measure as the weights. The fact that this method relies on no prior information is an advantage, but it also makes it difficult to find a generalization method that is both efficient and theoretically sound. As an alternative, a number of authors have proposed to use *class-based models*, in which the classes (similarities) present in an existing thesaurus are used for the purpose of smoothing estimated probabilities.

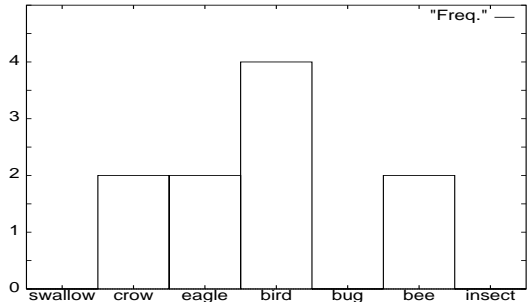

Figure 1: Frequency data for the subject slot of verb 'fly'

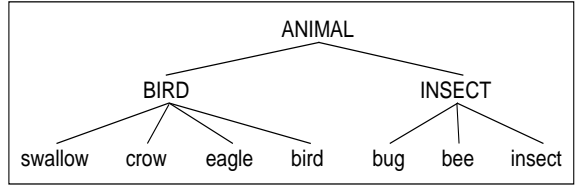

Figure 2: An example thesaurus

## 2.2 Class-Based Models

An example of a class-based method is Resnik's method of generalizing case slots using a thesaurus and the so-called *selectional association* measure. The selectional association $A(v, s, C)$ is defined as follows,

$$A(v,s,C) = P(C|v,s) \times \log \frac{P(C|v,s)}{P(C)} \qquad (2)$$

where $C$ is a class of nouns present in a given thesaurus, and $v, s$ are a verb and a slot name, as described earlier. In generalizing a given noun $n$ using this method, the noun class $C$ with the *maximum* $A(v, s, C)$, among all super classes of $n$ in a given thesaurus is output. This method is based on an interesting intuition, but its interpretation as a method of estimating probability distributions is yet to be determined. We propose a class-based generalization method whose performance as a method of estimation is guaranteed to be near optimal.

In this paper, we define the class-based model in the following way. A *class-based model* consists of a *partition* of the set of nouns $\mathbf{N}$, namely $\Gamma \subseteq 2^{\mathbf{N}}$ such that $\cup_{C_i \in \Gamma} C_i = \mathbf{N}$ and $\forall C_i, C_j \in \Gamma \; C_i \cap C_j = \emptyset$, and a number of parameters specifying the conditional probability for each $C$ in that partition, namely

$$P(C|v,s). \qquad (3)$$

Within a given class $C$, it is assumed that each noun is generated with equal probability, i.e.,

$$\forall n \in C : P(n|v,s) = \frac{1}{|C|} \times P(C|v,s) \qquad (4)$$

Note that this assumption is basically equivalent to the assumption made in other class and similarity based methods (Hindle 90; Grishman & Sterling 94) that similar words occur in the same context with roughly equal likelihood.

## 2.3 The Tree Cut Model

Mainly for the consideration of computational tractability, we reduce the number of possible partitions to consider by using an existing thesaurus as prior knowledge. That is, we restrict our attention on those partitions that exist within the thesaurus in the form of a *cut* in the tree. Here we mean by a 'thesaurus' a tree in which each leaf node stands for a noun, while each internal node represents a noun class, and domination stands for set inclusion. (See Figure 2.) A *cut* in a tree is any set of nodes in the tree that defines a partition of the leaf nodes, viewing each node as representing the set of all the leaf nodes it dominates. For example, in the thesaurus of Figure 2, there are five cuts: [ANIMAL],[BIRD, INSECT], [BIRD, bug, bee, insect], [swallow, crow, eagle, bird, INSECT], and [swallow, crow, eagle, bird, bug, bee, insect]. The class of *tree cut models* of a fixed thesaurus tree is then obtained by restricting the partition $\Gamma$ in the definition of a class-based model to be those that are present as a cut in that thesaurus tree. Formally, a tree cut model can be represented by a pair consisting of a tree cut, and a probability parameter vector of the same length.[3] For example, $M =$ ([BIRD, bug, bee, insect], [0.8, 0, 0.2, 0]) is a tree cut model,[4] which is shown in Figure 4.

---

[1] Since the case slots in a case frame are in general not independent, generalization of case frames involves generalization of individual case slots, and learning of the dependencies that exist between different case slots. In this paper we confine ourselves to the former problem of generalizing case slots. (We will address the latter issue in the near future.)

[2] A representation of a (conditional) probability distribution is usually called a *probability model*, or simply a *model*.

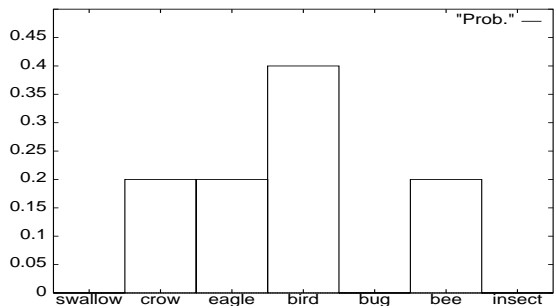

Figure 3: A tree cut model with [swallow,crow,eagle,bird,bug,bee,insect]

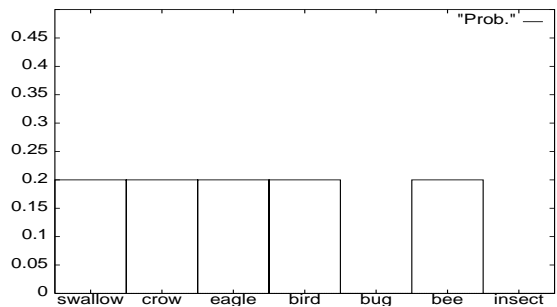

Figure 4: A tree cut model with [BIRD,bug,bee,insect]

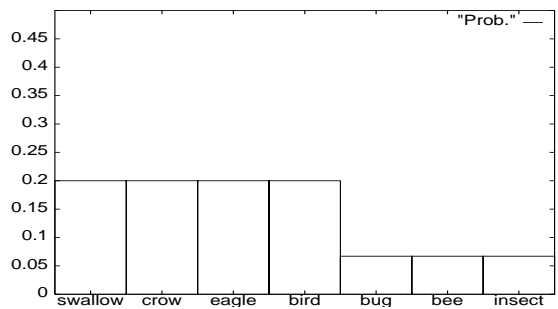

Figure 5: A tree cut model with [BIRD,INSECT]

Recall that $M$ defines a (conditional) probability distribution $P_M(n|v,s)$ in the following way: For any word that is in the tree cut, such as 'bee', the probability is given as explicitly specified by the model, i.e. $P_M(\text{bee}|\text{fly},\text{arg1}) = 0.2$. For any class in the tree cut, the probability is distributed uniformly to all words dominated by it. For example, since there are four words that fall under the class BIRD, and 'eagle' is one of them, $P_M(\text{eagle}|\text{fly},\text{arg1}) = 0.8/4 = 0.2$. Note that in this way, $M$ 'smoothes' the probabilities assigned to the nouns under BIRD, even if they may have different observed frequencies. If we use MLE for the parameter estimation, we can obtain five tree cut models, shown in Figures 3-5, from the co-occurrence data in Figure 1. We have thus formalized the problem of generalizing a case slot as that of estimating a model from the class of tree cut models for some fixed thesaurus tree, namely selecting a model which best explains the data from among the class of tree cut models.[5]

## 3 Generalization Method Based On MDL

As our estimation strategy, we employ the *MDL* (Minimum Description Length) principle (Rissanen 78; Rissanen 84; Rissanen 86). *MDL* is a principle of data compression and statistical estimation from information theory, which states that the best probability model for given data is that which requires the least code length in bits for the encoding of the model itself and the given data observed through it[6]. The former is called 'the model description length' and the latter 'the

---

[3] In general, a probability model consists of a discrete model and a probability parameter vector. The tree cut is the discrete model in this case. In the sequel, we sometimes use the discrete model (tree cut) to refer to a tree cut model, when the values of the probability parameters are clear from the context.

[4] Note that the probability parameters in $M$ were estimated using MLE from the co-occurrence data in Figure 1, i.e. $f(\text{BIRD}|\text{fly},\text{arg1}) = 8$, $f(\text{bee}|\text{fly},\text{arg1}) = 2, \text{others} = 0$.

[5] There have been a number of methods proposed in the literature, which will automatically construct the thesaurus itself using co-occurrence data(Hindle 90; Brown et al. 92; Pereira et al. 93). In this paper, we use an existing thesaurus for efficiency purpose, although we can extend (and we have extended) our method so as to automatically construct a thesaurus.

[6] We refer the interested reader to (Quinlan & Rivest 89) for an introduction to MDL principle.

data description length.'

In our current problem, it tends to be the case, in general, that a model near the root of the thesaurus tree, such as that in Figure 5, is simpler (in terms of the number of parameters), but tends to have a poorer fit to the data. In contrast, a model near the leaves of the thesaurus tree, such as that in Figure 3, is more complex, but tends to have a better fit to the data. In other words, there is a trade-off between the simplicity of a model and the goodness of fit to the data. While the model description length of MDL is an indicator of the former, the data description length is of the latter. MDL claims that the model which minimizes the sum total of the description lengths should be the best model.

In the remainder of this section, we will describe how we apply MDL to our current problem. We will then discuss the rationals behind using MDL in our present context.

### 3.1 Calculating the Description Length

We first show how the description length for a model is calculated. Given a tree cut model $M$ and data $S$, its total description length[7] $L(M)$ is computed as the sum of the model description length $L_{mod}(M) + L_{par}(M)$, and data description length $L_{dat}(M)$. Namely,

$$L(M) = L_{mod}(M) + L_{par}(M) + L_{dat}(M) \quad (5)$$

$L_{mod}(M)$ is calculated by

$$L_{mod}(M) = \log |\mathcal{G}| \quad (6)$$

where $\mathcal{G}$ denotes the set of cuts in the tree $T$. This is because if there are $|\mathcal{G}|$ cuts in the tree, then we need $\log |\mathcal{G}|$ bits to describe each of the cuts (for further explanation see (Quinlan & Rivest 89)). Throughout this paper 'log' denotes the logarithm to the base 2. $L_{par}(M)$, often referred to as the parameter description length, is calculated by,

$$L_{par}(M) = \frac{K}{2} \times \log |S| \quad (7)$$

where $K$ denotes the number of (free) parameters in the cut model, i.e. $K$ equals the number of nodes in $M$ minus one. It is known to be best to use $\log \sqrt{|S|} = \frac{\log |S|}{2}$ bits to describe each of the parameters.[8] Finally, $L_{dat}(M)$ is calculated by

$$L_{dat}(M) = -\sum_{n \in S} \log \hat{P}(n) \quad (8)$$

where for simplicity we write $\hat{P}(n)$ for $P_M(n|v,s)$ with the parameters estimated using the MLE estimate, which is equivalent to minimizing the data description length. (We will elaborate on why this is the case in Subsection 3.3.) Recall that $\hat{P}(n)$ is obtained by normalizing the frequencies, i.e.,

$$\forall n \in C, \hat{P}(n) = \frac{1}{|C|} \times \hat{P}(C) \quad (9)$$

$$\forall C \in \Gamma, \hat{P}(C) = \frac{f(C)}{|S|} \quad (10)$$

where $f(C)$ denotes the total frequency of nouns in class $C$ in sample $S$, and $\Gamma$ a cut.

---

[7]$L(M)$ depends on $S$, but we will leave $S$ implicit.

[8]One can interpret this as follows. The standard deviation of MLE is $O(\frac{1}{\sqrt{|S|}})$, and hence the precision required for each parameter is $O(\log \sqrt{|S|}) = O(\frac{\log |S|}{2})$.

With the description length of a tree cut model defined in the above manner, we wish to select a model with the minimum description length and output it as the result of generalization. Since we assume here that every cut has an equal $L_{mod}$, technically we need only calculate and compare $L'(M) = L_{par}(M) + L_{dat}(M)$ as the description length of a model. For simplicity, in the sequel we will sometimes write just $L'$ or $L'(\Gamma)$ for $L'(M)$, where $\Gamma$ is the tree cut of $M$.

The description lengths for the data in Figure 1 using various tree cut models of the thesaurus in Figure 2 are shown in Table 2. (Table 1 shows how the description length is calculated for the cut [BIRD,bug,bee,insect].) These figures indicate that the model in Figure 5 is the best model, according to MDL. (Note, as we will see in Subsection 3.2, that with different co-occurrence data, a different tree cut might be optimal.)

| $C$ | BIRD | bug | bee | insect |
|---|---|---|---|---|
| $f(C)$ | 8 | 0 | 2 | 0 |
| $|C|$ | 4 | 1 | 1 | 1 |
| $\hat{P}(C)$ | 0.8 | 0.0 | 0.2 | 0.0 |
| $\hat{P}(n)$ | 0.2 | 0.0 | 0.2 | 0.0 |
| cut | [BIRD,bug,bee,insect] | | | |
| $L_{par}$ | $\frac{(4-1)}{2} \times \log 10 = 4.98$ | | | |
| $L_{dat}$ | $-(2+4+2+2) \times \log 0.2 = 23.22$ | | | |

Table 1: Parameters in the model of cut [BIRD,bug,bee,insect]

### 3.2 An Efficient Algorithm

In generalizing a case slot using MDL, we could in principle calculate the description length of every possible model and output a model with the minimum description length as the generalization result, if computation time were of no concern. But since the number of cuts in a thesaurus tree of noun is exponential (for example, for a complete $b$-ary tree of depth $d$ it is of order $O(2^{b^{d-1}})$), it is impractical to do so. Nonetheless, we were able to devise a simple and efficient algorithm, which is guaranteed to find a model with the minimum description length.

Our algorithm, which we call Find-MDL, recursively finds the optimal MDL model for each child subtree of a given node and appends all the optimal models of these subtrees and returns it, unless collapsing all the lower-level optimal models into a single node (that is, a single class) reduces the description length, in which case it does do so. The details of the algorithm are given below. Note that for simplicity we describe Find-MDL as outputting a cut, rather than a complete model. (It is implicitly assumed that the parameters are estimated using MLE.)

| Tree cut model | $L_{par}$ | $L_{dat}$ | $L'$ |
|---|---|---|---|
| [ANIMAL] | 0 | 28.07 | 28.07 |
| [BIRD,INSECT] | 1.66 | 26.39 | 28.05 |
| [BIRD,bug,bee,insect] | 4.98 | 23.22 | 28.20 |
| [swallow,crow,eagle,bird,INSECT] | 6.64 | 22.39 | 29.03 |
| [swallow,crow,eagel,bird,bug,bee,insect] | 9.97 | 19.22 | 29.19 |

Table 2: Description lengths of the tree cut models

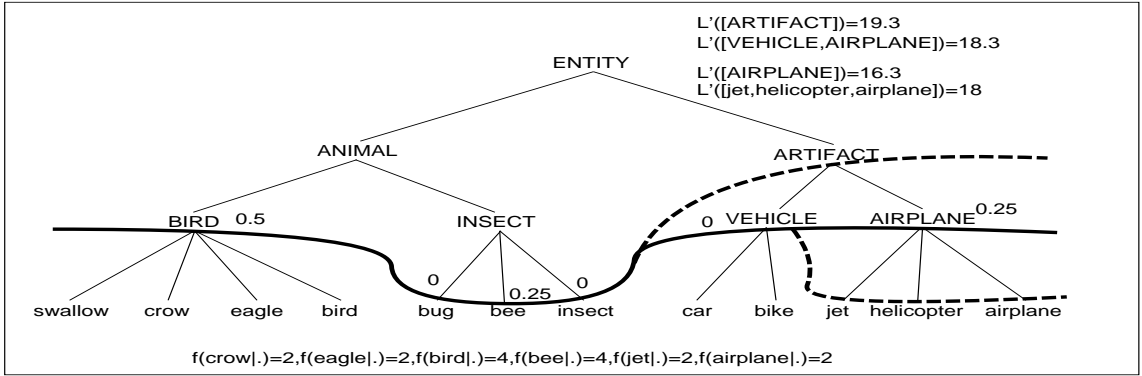

Figure 6: An example application of Find-MDL

Here we let $t$ denote a thesaurus tree,
root($t$) the root of the tree.
Initially $t$ is set to the entire tree.
**algorithm** Find-MDL($t$) := cut
1.    **if**
2.       $t$ is a leaf node
3.    **then**
4.       return([$t$])
5.    **else**
6.       For each child tree $t_i$ of $t$
7.       $c_i$ :=Find-MDL($t_i$)
8.       $c$ := append($c_i$)
9.       **if**
10.          $L'([\text{root}(t)]) < L'(c)$
11.       **then**
12.          return([root($t$)])
13.       **else**
14.          return($c$)

Figure 7: The algorithm: Find-MDL

Note in the above algorithm that the parameter description length is calculated as $\frac{K+1}{2} \log |S|$, where $K+1$ is the number of nodes in the current cut, both when $t$ is the entire tree or when it is a proper subtree. This contrasts with the fact that the number of *free* parameters is $K$ for the former, while it is $K+1$ for the latter. For the purpose of finding a tree cut with the minimum description length, however, this distinction can be ignored. (c.f. Appendix)

Figure 6 illustrates how the algorithm works: In the recursive application of Find-MDL on the subtree rooted at AIRPLANE, the if-clause on line 10 evaluates to true since $L'([\text{AIRPLANE}]) = 16.3$ and $L'([\text{jet}, \text{helicopter}, \text{airplane}]) = 18$, and hence [AIRPLANE] is returned. Then in the call to Find-MDL on the subtree rooted at ARTIFACT, the same if-clause evaluates to false since $L'([\text{ARTIFACT}]) = 19.3$ and $L'([\text{VEHICLE}, \text{AIRPLANE}]) = 18.3$, and hence [VEHICLE, AIRPLANE] is returned.

Concerning the above algorithm, we show that the following proposition holds, whose proof can be found in Appendix A.

**Proposition 1** *The algorithm Find-MDL terminates in time $O(N \times |S|)$, where $N$ denotes the number of leaf nodes in the input thesaurus $T$ and $|S|$ denotes the input sample size, and outputs a tree cut model of $T$ with the minimum description length.*

### 3.3 Estimation, Generalization and MDL

When a discrete model (a partition $\Gamma$ of the set of nouns $\mathcal{N}$ in our present context) is fixed, and the estimation problem involves only the estimation of probability parameters, the classic maximum likelihood estimation (MLE) is known to be satisfactory. In particular, the estimation of a word-based model is one such problem, since the partition is fixed and equals $\mathcal{N}$. Furthermore, for a fixed discrete model, it is known that the MLE coincides with MDL: Given data $S = \{x_i : i = 1, ..., m\}$, the MLE estimate $\hat{P}$ maximizes the 'likelihood' of the data, that is,

$$\hat{P} = \arg\max_P \prod_{i=1}^m P(x_i). \quad (11)$$

It is easy to see that $\hat{P}$ also satisfies

$$\hat{P} = \arg\min_P \sum_{i=1}^m -\log P(x_i). \quad (12)$$

This is nothing but the MDL estimate in this case, since $\sum_{i=1}^m -\log P(x_i)$ is the *data description length*.

When the estimation problem involves model selection, i.e. the choice of a tree cut in the present context, MDL's behavior significantly deviates from that of MLE. This is because MDL insists on minimizing the sum total of the data description length *and* the model description length, while MLE is still equivalent to minimizing the data description length only. So for our problem of estimating tree cut models, MDL tends to select a cut that is reasonably simple yet fits the data quite well, whereas the model selected by MLE will be a *word-based model*, as it will always manage to fit the data as well as any tree cut model.

In statistical terms, the superiority of MDL as an estimation method is related to the fact which we noted earlier that even though a word-based model can provide the best fit to the given data, the estimation of the parameters are poor as there are too many parameters to estimate. So MLE, when applied on a data set of a modest size, is likely to estimate most parameters as zero, and thus suffers from the data sparseness problem. Note in Table 2, that MDL avoids this problem by taking into account the model complexity as well as the fit to the data.

MDL stipulates that the model with the minimum description length should be selected both for data compression and estimation. This intimate connection between estimation and data compression can also be thought of as that between estimation and *generalization*, since in order to compress information, there needs to be generalization. In our current problem, this corresponds to the generalization of individual nouns present in case frame instances in the data as classes of nouns present in a given thesaurus. For example, given the thesaurus in Figure 2 and frequency data in Figure 1, we would like our system to judge that the class 'BIRD' and the word 'bee' can be the subject of the verb 'fly.' The problem of deciding whether to stop generalizing at 'BIRD' and 'bee,' or generalizing further to 'ANIMAL' has been addressed by a number of authors (c.f. (Resnik 92; Resnik 93)). Minimization of the total description length provides a disciplined criterion to do this.

The remarkable fact about MDL is that theoretical findings (c.f. (Barron & Cover 92; Yamanishi 92)) have indeed verified that MDL, as an estimation method, is near optimal,[9] in terms of the speed of convergence[10] of its estimated models to the *true* model, as the data size increases. Thus MDL provides (i) a way of smoothing probability parameters to solve the data sparseness problem, and at the same time (ii) a way of generalizing nouns in the data to noun classes of an appropriate level, both *as a corollary* to the near optimal *estimation* of the distribution of the input data.

A frequently asked question in cognitive science is that of why humans learn, and it is believed by many that there are two major motivations: To improve the accessibility of accumulated knowledge, and to interpret new information (Rumelhart & Norman 78). The fact that MDL is suited for both compression and estimation seems to be an affirmative evidence for MDL as a possible cognitive model. For example, our method of generalizing case frames based on MDL will output a compact representation summarizing the observed data, which is also near optimal for predicting the acceptability of unseen instances in the future. Thus we feel that our method is not only mathematically sound, but also cognitive scientifically well-motivated.

## 4 Experimental Results

### 4.1 Experiment 1

First, we extracted *head, slot_name, slot_value* triples from the texts of the *tagged* Wall Street Journal corpus (ACL/DCI CD-ROM1) consisting of 126,084 sentences, using conventional pattern matching techniques, then applied the algorithm Find-MDL to generalize the *slot_value*s of the triples.

When generalizing, we used the noun taxonomy of WordNet (version1.4) (Miller et al. 93) as our thesaurus. The noun taxonomy of WordNet has a structure of DAG and the (leaf and internal) nodes stand for a word sense and not a word, and thus often contain several words of the same word sense. Since it does not meet the assumption we made on our thesaurus, we used it in the following modified form. First, the observed frequency of a word as a *slot_value* of given *head* and *slot_name* is equally divided be-tween all the nodes containing that word. Then the frequency of an internal node is calculated as the sum of the frequencies of all the nodes it dominates. Finally we applied our generalization algorithm to the tree obtained by discarding from the thesaurus those subtrees that are rooted at any node containing a word that actually occurs as the *slot_value*.

---

[9] There is a Bayesian interpretation of MDL: MDL is essentially equivalent to *the posterior mode* in the Bayesian terminology. It is known that in fact, *Bayesian posterior mixture* is optimal in some sense, but it is also known that in many cases these two estimates are approximations of each other (Takeuchi 95).

[10] The models selected by MDL converge to the true model approximately at the rate of $1/K^*$ where $K^*$ is the number of parameters in the true tree cut model, where as for MLE the rate is $1/N$, where $N$ is the number of leaf nodes.

| direct object of 'eat' | | |
|---|---|---|
| Class | Prob. | Example words |
| ⟨food,nutrient⟩ | 0.39 | pizza |
| ⟨life_form,organism,being,living_thing⟩ | 0.11 | lobster |
| ⟨measure,quantity,amount,quantum⟩ | 0.10 | amount *of* |
| ⟨artefact,article,artefact⟩ | 0.08 | *as if eat* rope |
| direct object of 'buy' | | |
| Class | Prob. | Example words |
| ⟨object,inanimate_object,physical_object⟩ | 0.30 | computer |
| ⟨asset⟩ | 0.10 | stock |
| ⟨group,grouping⟩ | 0.07 | company |
| ⟨legal_document,legal_instrument,official_document,instrument⟩ | 0.05 | security |

Table 3: An example of generalization result

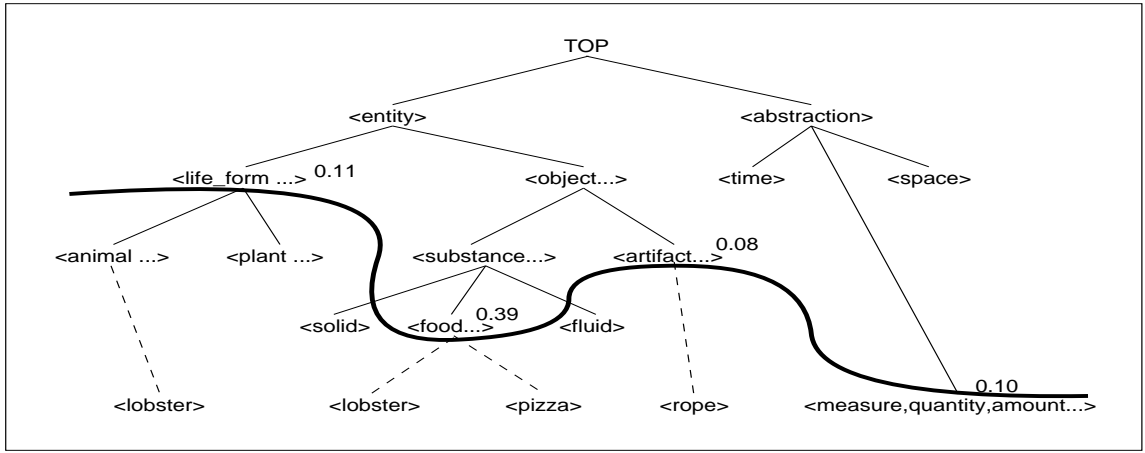

Figure 8: An example generalization result

Table 3 shows an example generalization result; for the direct object slot of the verb 'eat' and the verb 'buy', where ⟨...⟩ denotes a node in WordNet. Classes with probabilities less than the threshold of 0.05 are discarded. Figure 8 shows the corresponding cut in WordNet for the direct object slot of 'eat'. Note, for example, that the fact that ⟨animal⟩, ⟨plant⟩ were generalized to ⟨life_form⟩ seems reasonable because both of these categories are suited for the direct object slot of 'eat.' On the contrary, ⟨food⟩ was not generalized to ⟨substance⟩, which also seems correct because not all substances are edible. Thus, despite the fact that the employed extraction method is not noise-free, and word sense ambiguities remain after extraction, the generalization result seems to agree with our intuition to a satisfactory degree. This is probably because the 'noisy' part usually has a small probability and thus tends to be discarded. This, we believe, is another desirable consequence of using MDL as our estimation method. Now, since we can tag a plain text with a high accuracy with current technology (c.f. (Church 88)), we can acquire case frame patterns completely automatically using our generalization method, and thereby provide useful usage descriptions to lexicographers.

### 4.2 Experiment 2

We conducted another experiment in which we used the acquired knowledge to resolve pp-attachment ambiguities. First we selected about 10% of the parsed trees from the parsed Wall Street Journal corpus (Penn Tree Bank 1) as test data, and used the remainder as training data. Then we extracted 181,250 case frames from the training data using heuristics and extracted 172 ($verb, noun_1, prep, noun_2$) patterns from the test data. We generalized the *slot_value*s of the *head, slot_name, slot_value* triples using our method (Find-MDL algorithm and WordNet were used in the same manner as in experiment 1, with the output threshold set to 0.05) and selectional association based on Resnik's method. We then used them to disambiguate the 172 patterns. We also used lexical association proposed by Hindle & Rooth (Hindle & Rooth 91) to disambiguate those patterns. Although it would be possible to resolve word sense ambiguities as well, we confined ourselves to the structural disambiguation problem at this stage.

When using our method for disambiguation, we compare $\hat{P}(noun_2 | verb, prep)$ and $\hat{P}(noun_2 | noun_1, prep)$ to determine the attachment site of $(prep, noun_2)$. If the former is larger than the latter, we attach it to $verb$, else if the latter is larger than the former, we attach it to $noun_1$, and otherwise (especially when both are 0), we conclude that we cannot make a decision. Determining the attachment site in this way is natural and we empirically found that this gives us the best results in terms of accuracy. When using the selectional association to disambiguate, we heuristically calculate the 't-score' of $\max(A(Class_2 | verb, prep))$ and $\max(A(Class_2 | noun_1, prep))$, where the maximization is over $noun_2 \in Class_2$. If the t-score is not significant (at significance level 95%), we conclude that we cannot make a decision. When using the lexical association to disambiguate, we calculate the t-score of $\hat{P}(prep | verb)$ and $\hat{P}(prep | noun_1)$. Again if the t-score is not significant, we conclude that we cannot make a decision.

|         | Coverage(%) | Accuracy(%) |
|---------|-------------|-------------|
| Default | 100         | 70.2        |
| LA      | 87.2        | 86.0        |
| MDL     | 49.4        | 88.2        |
| SA      | 49.4        | 84.7        |
| MDL2    | 65.7        | 85.8        |

Table 4: Results of PP-attachment disambiguation

Table 4 shows the results of pp-attachment disambiguation in terms of 'coverage' and 'accuracy.' Here 'coverage' refers to the proportion (in percentage) of the test patterns on which the disambiguation method could make a decision. 'Default' refers to the method of always attaching $(prep, noun_2)$ to $noun_1$, while 'MDL,' 'SA,' and 'LA,' stand for using MDL, selectional association, and lexical association, respectively.

Here are some points that are worth noting about these results. First, although the coverage of LA is larger than those of both MDL and SA[11], we believe that this is mainly because it uses a model not as rish as those of MDL and SA, and thus needs less data to estimate its parameters. However, as Resnik correctly pointed out, if we hope to improve the performance of disambiguation as we get larger data sizes, we need a richer model such as those used in MDL and SA.

[11] Our result on LA is close to Hindle's, but deviates from Resnik's, probablely because of the different data used.

Second, the accuracy of MDL is better than that of SA, while its coverage is the same as that of SA. MDL tends to generalize only when there is enough evidence, and when it does, the result seems to fit the human intuition quit well. Table 5 shows an example generalization result for the 'on' slot for the verb 'watch'. Note that MDL does not generalize 'afternoon' because of its small frequency, while it has been generalized to 'acknowledgement' by SA, which seems rather odd.[12]

| Input | Freq. |
|-------|-------|
| watch on afternoon | 1 |
| watch on screen | 1 |
| watch on set | 2 |
| watch on street | 1 |
| watch on television | 2 |
| watch on tv | 2 |
| Output of MDL | Prob. |
| watch on ⟨entity⟩ | 0.59 |
| Output of SA | SA |
| watch on ⟨television,...⟩ | 1.78 |
| watch on ⟨artifact,...⟩ | 1.43 |
| watch on ⟨acknowledgment,...⟩ | 0.81 |
| watch on ⟨afternoon⟩ | 0.50 |

Table 5: A example generalization result

We also conducted the following additional experiment. We randomly selected 50%, 60%, 70%, 80%, and 90% of the training data and applied MDL and SA to them, repeated this process ten times, and then evaluated the accuracy and coverage, averaged over the ten trials. Figures 9 and 10 show the results of this experiment. We found that MDL outperforms SA throughout in terms of accuracy, and its coverage improves faster than SA as the data size increases.

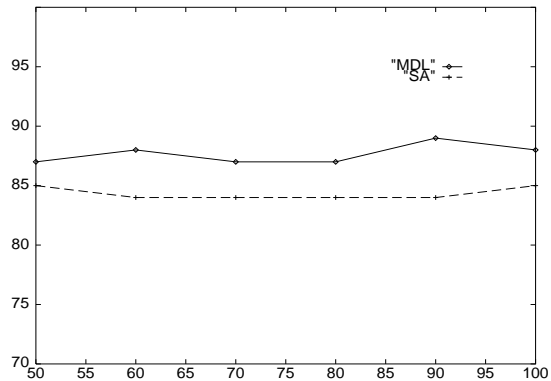

Figure 9: Accuracy of MDL and SA

[12] Note that 'afternoon' does not belong to '⟨entity⟩,' and that (some word-sense of it) lies within '⟨acknowledgement⟩,' as in 'good afternoon.'

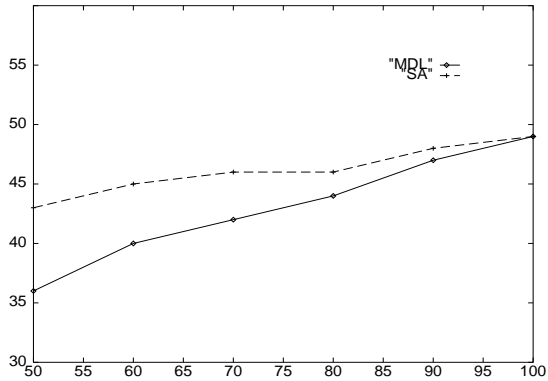

Figure 10: Coverage of MDL and SA

Admittedly, the coverage of MDL (and of SA) is not satisfactory. So we conducted another additional experiment, in which we also generalized the *head* of the triples in the data, provided *head* is a noun. When disambiguating, we compared $\hat{P}(noun_2|verb,prep)$ and $\hat{P}(noun_2|noun_1,prep)$. The result of this additional experiment is shown in Table 4 as MDL2, which indicates that the coverage can be significantly improved this way, although the accuracy drops somewhat. In fact, the method[13] used in the actual experiment in (Resnik 92) employs the version of SA in which both *slot_value* and *head* are generalized (provided *head* is a noun). This seems to have the effect of improving the coverage, since the reported coverage (67.2%) is better than that for the version of SA used here which generalized *slot_value* only, while the accuracy (84.6%) is the same as what we obtained.

Finally, we tested the method 'Combined', which applies MDL first, and then applies LA to the rest, and then finally uses Default on what remains. We also tested 'Combined2', which applies MDL2 first, and then LA, and finally Default. Table 6 shows the results of this experiment. Our final accuracy (84.9%) is better than that (78.3%) reported in (Hindle & Rooth 91) and that (82.2%) reported in (Resnik 92). We conclude that our method improves upon the existing methods, although its statistical significance is moderate. (The standard deviations for these three figures are 2.7% 1.4% and 2.9%, respectively.)

|  | Coverage(%) | Accuracy(%) |
| --- | --- | --- |
| Combined | 100 | 84.3 |
| Combined2 | 100 | 84.9 |

Table 6: Final Results of PP-attachment disambiguation

---

[13] We did not implement the exact method actually used in (Resnik 92).

## 5 Conclusions

We proposed a new method of generalizing case frames. We believe that our method has the following merits: (1) It is theoretically sound; (2) It is cognitive scientifically well-motivated; (3) It is computationally efficient; (4) It is robust against noise. The disadvantage of our method is that its performance depends on the structure of the particular thesaurus used. This, however, is a problem commonly shared by any generalization method which uses a thesaurus as prior knowledge. Our experimental results indicate that the performance of our method is better or at least as good as existing methods.

## Acknowledgement

We thank Mr. K. Nakamura and Mr. T. Fujita of NEC C&C Research Laboratories for their encouragement. We acknowledge the A.C.L. for providing the ACL/DCI CD-ROM, L.D.C. of the University of Pennsylvania for providing the Penn Tree Bank corpus data, and Princeton University for providing WordNet.

## A  Proof of Proposition 1

For an arbitrary subtree $T'$ of a thesaurus tree $T$ and an arbitrary tree cut model $M$ of $T$, let $M \cap T'$ denote the submodel of $M$ that is contained in $T'$. Also for any sample $S$ and any subtree $T'$, let $S \cap T'$ denote the subsample of $S$ contained in $T'$. Then define[14] $L_{dat}(M', S')$ to be the data description length of (sub)sample $S'$ using (sub)model $M'$, $L_{par}(M', |S|)$ the parameter description length for the parameters in (sub)model $M'$ with (total) sample size $|S|$, and finally $L'(M', S', |S|) = L_{dat}(M', S') + L_{par}(M', |S|)$ in general for any (sub)model $M'$ and (sub)sample $S'$ of $S$.

First note that for any (sub)tree $T$, (sub)model $M \cap T$, (sub)sample $S \cap T$, and $T$'s child subtrees $T_i : i = 1, .., k$, we have

$$L_{dat}(M \cap T, S \cap T) = \sum_{i=1,..,k} L_{dat}(M \cap T_i, S \cap T_i). \quad (13)$$

This follows from the mutual disjointness of the $T_i$, and the independence of the parameters in the $T_i$. We also have, when $T$ is a *proper* subtree of the entire thesaurus tree,

$$L_{par}(M \cap T, |S|) = \sum_{i=1,..,k} L_{par}(M \cap T_i, |S|). \quad (14)$$

Since the number of free parameters of a model in the entire thesaurus tree equals the number of nodes in the model *minus* one due to the stochastic condition (that the probability parameters must sum to one), when $T$ equals the entire thesaurus tree, theoretically the parameter description length for a tree cut model of $T$ should be

$$L_{par}(M \cap T, |S|) = \sum_{i=1,..,k} L_{par}(M \cap T_i, |S|) - \frac{\log |S|}{2}. \quad (15)$$

Since the second term $-\frac{\log |S|}{2}$ in (15) is constant once the input sample $S$ is fixed, for the purpose of finding a model with the minimum description length, it is irrelevant. We will thus use the identity (14) both when $T$ is the entire tree and when it is a proper subtree. (This allows us to use the same recursive algorithm (Find-MDL) in all cases.)

It follows from (13), and (14) that the minimization of description length can be done essentially independently for each subtree. Namely, if we let $L'_{opt}(M \cap T, S \cap T, |S|)$ denote the minimum description length achievable for the (sub)model $M \cap T$ on the (sub)sample $S \cap T$, $\hat{P}_S(\eta)$ the MLE estimate for node $\eta$ using sample $S$, and root$(T)$ the root node of (sub)tree $T$, then we have

$$L'_{opt}(M \cap T, S \cap T, |S|) = \\ \min\{L'(([\text{root}(T)], [\hat{P}_S(\text{root}(T))]), S \cap T, |S|), \\ \sum_{i=1,..,k} L'_{opt}(M \cap T_i, S \cap T_i, |S|)\}. \quad (16)$$

---

[14] Note that in Section 3 $L_{dat}$, $L_{par}$ and $L'$ were defined as functions of one argument, leaving the dependency on the sample implicit. Here we make it explicit as the sample does not always equal $S$.

The rest of the proof proceeds by induction. First, when $T$ consists of a single leaf node, the MLE for the class represented by $T$ is returned, which is known to minimize the data description length. (Clearly, the parameter description length is identical for all.) Next, inductively assume that Find-MDL($T'$) correctly outputs a model with the minimum description length for any tree $T'$ of size less than $n$. Then, given a tree $T$ of size $n$ whose root node has at least two children, say $T_i : i = 1, .., k$, for each $T_i$, Find-MDL($T_i$) returns a model with the minimum description length by the inductive hypothesis. Then, since (16) holds, whichever way the if-clause on lines 9, 10 of Find-MDL evaluates to, what is returned on line 12 or line 14 will still be a model with the minimum description length, completing the inductive step. It is easy to see that the running time of the algorithm is linear in both the size of the input thesaurus tree and the sample size. □